\documentclass{sig-alternate-05-2015}
	\usepackage{enumitem}
	\usepackage{graphicx}
	\usepackage{times}
	\usepackage{balance}
	\usepackage{latexsym}
	\usepackage{graphicx}
	\usepackage{amssymb,amsmath}
	\newcommand{\suchthat}{\;\ifnum\currentgrouptype=16 \middle\fi|\;}
	\usepackage{lingmacros}
	\usepackage{tree-dvips}
	\usepackage{amssymb}
	\usepackage{amsmath}
	\usepackage{graphicx}
	\usepackage{float} %pitanje
	\usepackage{graphicx}
	\usepackage[font=small,skip=0pt]{caption}
	\usepackage{wrapfig}
	\usepackage{booktabs}
	\usepackage{array}
	\usepackage{indentfirst}
	\usepackage{caption}
	\usepackage{subcaption}
	\usepackage[multiple]{footmisc}
	
	\usepackage{multirow}
	\usepackage[table]{xcolor}
	\definecolor{lightgray}{gray}{0.85}
	
	\usepackage{algorithm}
	\usepackage{algpseudocode}
	\usepackage{pifont}
\newcolumntype{C}[1]{>{\centering\arraybackslash}p{#1}}

\newcommand\numberthis{\addtocounter{equation}{1}\tag{\theequation}}

\begin{document}

\CopyrightYear{2016} 
\setcopyright{acmcopyright}
\conferenceinfo{CIKM'16 ,}{October 24-28, 2016, Indianapolis, IN, USA}
\isbn{978-1-4503-4073-1/16/10}\acmPrice{\$15.00}
\doi{http://dx.doi.org/10.1145/2983323.2983854}

\title{Modeling Customer Engagement from Partial Observations}
	
%	Structured Regression on Deficient Data for Predicting Customer Engagement}

\numberofauthors{1} 
\author{
	\alignauthor
	Jelena Stojanovic, Djordje Gligorijevic, Zoran Obradovic \\
	%\titlenote{Anonymous Author}\\
	\affaddr{Computer \& Information Sciences Department, Temple University}\\
	       \affaddr{1925 N 12th Street}\\
	       \affaddr{Philadelphia, USA}\\
	\email{\{jelena.stojanovic, gligorijevic, zoran.obradovic\}@temple.edu}
}

\maketitle
\begin{abstract}
It is of high interest for a company to identify  customers expected to bring the largest profit in the upcoming period. Knowing as much as possible about each customer is crucial for such predictions. However, their demographic data, preferences, and other information that might be useful for building loyalty programs is often missing. Additionally, modeling relations among different customers as a network can be beneficial for predictions at an individual level, as similar customers tend to have similar purchasing patterns. We address this problem by proposing a robust framework for structured regression on deficient data in evolving networks with a supervised representation learning based on neural features embedding. The new method is compared to several unstructured and structured alternatives for predicting customer behavior (e.g. purchasing frequency and customer ticket) on user networks generated from customer databases of two companies from different industries. The obtained results show $4\%$ to $130\%$ improvement in accuracy over alternatives when all customer information is known. Additionally, the robustness of our method is demonstrated when up to $80\%$ of demographic information was missing where it was up to several folds more accurate as compared to alternatives that are either ignoring cases with missing values or learn their feature representation in an unsupervised manner.
\end{abstract}

%
% The code below should be generated by the tool at
% http://dl.acm.org/ccs.cfm
% Please copy and paste the code instead of the example below. 
%


%
% End generated code
%

%
%  Use this command to print the description
%


\section{Introduction}

Companies utilize loyalty programs to enforce personalized customer relationship management. 
These programs can be considered as a guiding force of marketing endeavors, as good loyalty program has the power to turn a business into a customer-oriented profit machine \cite{yi2003effects}. 
Users' behavior can greatly differ, and so rewards and promotions that do not care about each individual user's behavior can result in a severe revenue decline \cite{stauss2005customer}.

In order to wisely plan enhancement of future customer engagement, it could be of great use to predict future behavior of customers including their customer's ticket\footnote{Customer ticket is a common term for total dollar amount of transactions that customer spend over a certain period of time} and visit frequency, which are good indicators of purchasing habits, and are also important indicators of the company's success. Therefore, companies are interested to detect different types of customers and to model their purchasing habits in order to properly build customer-tailored loyalty programs \cite{dowling1997customer} and deepen customers loyalty to the brand.

In order to model users' behavior, a large variety of data needs to be collected. Each user action generates plenty of useful information about the user's habits, rendering loyalty programs one of marketing's biggest data-generating mechanisms\footnote{http://data-informed.com/customers-view-loyalty-programs-caution/ accessed May 2016}. Collection of data on actions and purchase details of customers can be fairly easy, but collection of demographic and other preference data may be challenging.
Even though users are usually willing to turn over their basic demographic data (such as gender or date of birth) in exchange for perceived value, they are often dissuaded from using loyalty services if required  to provide more than basic information or answer questionnaires. Additionally, many users may completely skip providing any demographic information if it is not obligatory. As such, analyzing customer data often requires dealing with a large fraction of missing values, which can severely limit the representational power of predictive models. However, companies would still like to infer about their customers in order to identify where customer-engagement marketing efforts should take place. For instance, based on certain estimates, such as forthcoming amount spent or visit frequency, they can quickly react to the market demand using personal recommendations via both online and offline channels, and/or by setting special offers with rewards, and discounts in order to increase the rate of customer retention.
\begin{figure*}[htb!]
	\centering
	\includegraphics[width=0.95\textwidth]{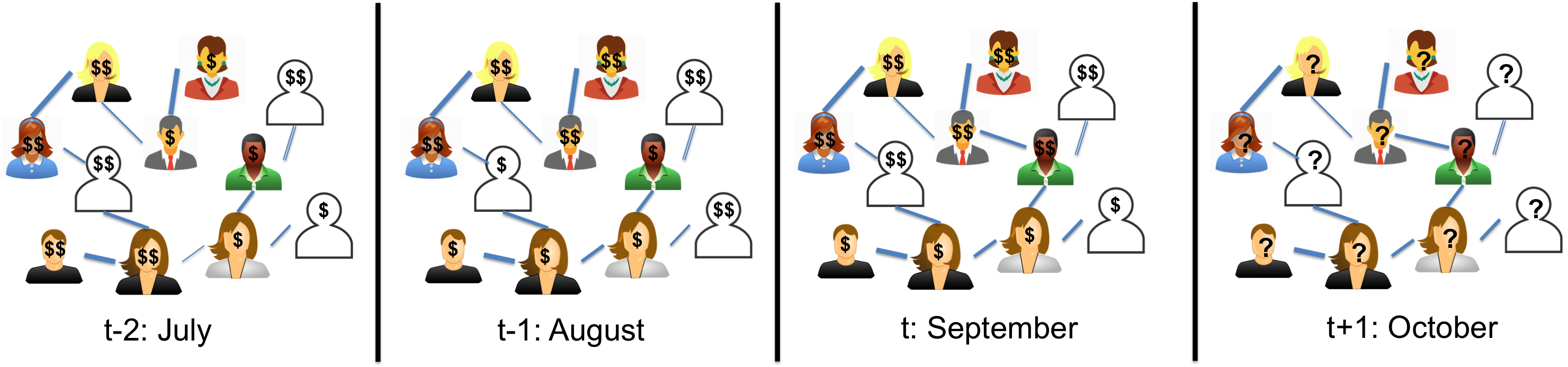}
	\caption{\textbf{Attributed} (purchase history and demographic data) \textbf{weighted} (different strength of customers similarity) \textbf{temporal} (months/quarters) \textbf{partially observed} (missing data in some of the nodes) \textbf{network} of customers %\footnotemark
		in which explanatory variables ($X$) are partially observed (blank users: demographic data are missing) and the response variables ($y$) represent measurements of customer engagement (customer ticket or visit frequency). The goal is to learn parameters of the model on training data ($..., t-2, t-1, t$) and predict continuous responses on test examples ($t+1$).}
	\label{graph}
\end{figure*}

A valuable source of information can be found in latent relations of customers.
As similar customers tend to have similar purchasing behavior, aforementioned predictive objectives can potentially be achieved by modeling those relations and observing customers as nodes in a network. Therefore, our focus is primarily set on the structured models that are capable of utilizing such information. 
Continuous (Gaussian) Conditional Random Fields are such a model developed for structured regression \cite{Radosavljevic2010}, that has been successfully applied to a large variety of domains including climate \cite{Radosavljevic2014,Stojanovic2015}, energy \cite{Kezunovic2016}, social networks \cite{uversky2014panning} and healthcare \cite{Gligorijevic2015,Gligorijevic2016}. 

This model is capable of structured regression for predicting customer tickets and visit frequencies, while modeling relationships among customers. However, it is limited to a given representation of the data, and is not robust to deficiency in explanatory variables. In order to improve representational and predictive power of this model, as well as to provide model robustness when a large fraction of explanatory variables are missing,  we propose a supervised neural based feature embedding approach capable of determining a latent feature representation from partially observed explanatory variables within a structured regression framework. 

The key contributions of this paper are summarized below:
\begin{itemize}
	\item Modeling customer data is formulated as a structured regression problem, with emphasis on prediction of future customer's ticket and visit frequency, where a novel deep structured feature learning framework is proposed for joint learning of customers representation and their correlations in a supervised manner;
	\item The robustness of the approach is demonstrated while missing a large fraction of very useful demographic data in various patterns on several tasks (up to 80\% of missing values);
	\item The model has shown experimental benefits compared to ten alternative models, including ones that are ignoring cases with scarce demographics as well as those that try to compensate for the deficiency of demographic data in an unsupervised fashion;
	\item The power and the generalization ability of the proposed approach are demonstrated on two challenging customer engagement applications on real-life data from different industries.
\end{itemize}

\section{Data}
\label{sec:data}
Customer engagement problems and proprietary datasets used to characterize effectiveness of the proposed method versus alternatives are described in this section.

\subsection{Customer engagement data} \label{sec:clutch_data}
Data from the business domain used in this study are based on electronically collected customer engagement information. Besides their purchase history (e.g. number of visits, items bought, discounts used, spending, etc.), we are partially familiar with their demographics, such as gender, age and similar information that a customer is asked to provide during an online registration/enrollment process.
However, as previously mentioned, there is a number of reasons why customers would not provide their demographics. Furthermore, a company can decide to simplify enrollment process for the convenience of customers, thus choosing not to collect a valuable set of information. Even though some informative data about customers is missing, we would still like to accurately infer their future spending habits and the frequency of their visits. 

For these two problems, in our experiments we drew datasets from two companies involved in different industry domains:
\begin{itemize}
	\item The first company is from the entertainment industry and a large part of their loyalty programs are based on the monthly membership fee, thus it is important to estimate how often a customer will visit in the following month. 
	\item The second company represents a global luxury lifestyle brand in the body and home products industry, which bases their members' rewards on quarterly spending. Estimating how much different customers will spend in the next quarter can be used to come up with new exclusive special offers and rewards to deepen customer engagement with the brand.
\end{itemize}

Therefore, we conducted experiments reported in Section~\ref{sec:experimental_evaluation} aimed to account for predictions of customers' future spending and visit frequency based on their recorded purchase history and their partially observed demographics. In the first application, data is collected over several months. However, the number of members for which demographics are known measures in the order of thousands. In the second application data is collected from the year 2012 and is aggregated on a quarterly level. The number of customers from the sample used in our experiments for which we know complete information (so that we can examine the influence of different processes and control experiments) measures in the order of hundreds.

Identification of the two companies is not shown in this study for privacy reasons. Also, showing sensitive information such as exact numbers of customers or exact customers' tickets and visit frequencies from the companies' databases is not being reported in this study. 

\subsection{Problem set-up}
\label{sec:problem_setup}

For each company we have a set of $N$ customers $c_i \in \mathcal{C}= \{c_1, ..., c_N\}$. For each customer $c_i$ we are familiar with the response variable $y_i$ and a vector of $m$ explanatory attributes: a $P$-dimensional vector of purchase data collected by the transaction system $x_{p}$ %=[x_{p_1},..., x_{p_P}]$ 
and a $D$-dimensional vector of (partially) observed demographic data $x_{d}$%=[x_{d_1}, ..., x_{d_D}]$
, so that $m=P+D$ and $x_i= [x_p^{(i)}, x_d^{(i)}]$.

\begin{figure}[h!]
	\centering
	\includegraphics[width=0.45\textwidth]{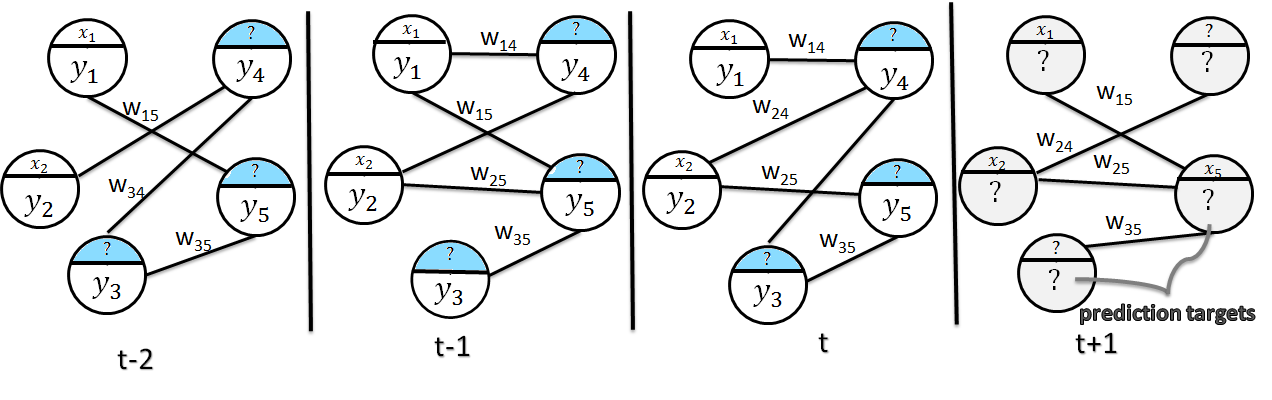}
	\caption{Attributed weighted network of customers observed over time in which explanatory variables ($X$) are partially missing (blue nodes) and dependent variables ($y$) represent the measure of customer engagement. The goal is to learn parameters of the model on data of the initial $t$ steps and predict continuous response values $y$ at time step $t+1$ (grey nodes).}
	\label{fig:graph}
\end{figure}

We observe a set of $N$ customers $\mathcal{C}$ over $T$ time steps and model them as a network as shown in Figure~\ref{fig:graph}. Edges in the graph are weighted and represent similarity of response variables of the nodes. The goal is to predict values of the response variable $y$ in each node of the graph in the following time step $t+1$.
\section{Related Work} \label{sec:related_work}
\begin{figure}[ht!]%{0.35\textwidth} %[11]
	\centering
	\begin{subfigure}{.33\textwidth}
		\begin{center}
			\includegraphics[width=\linewidth]{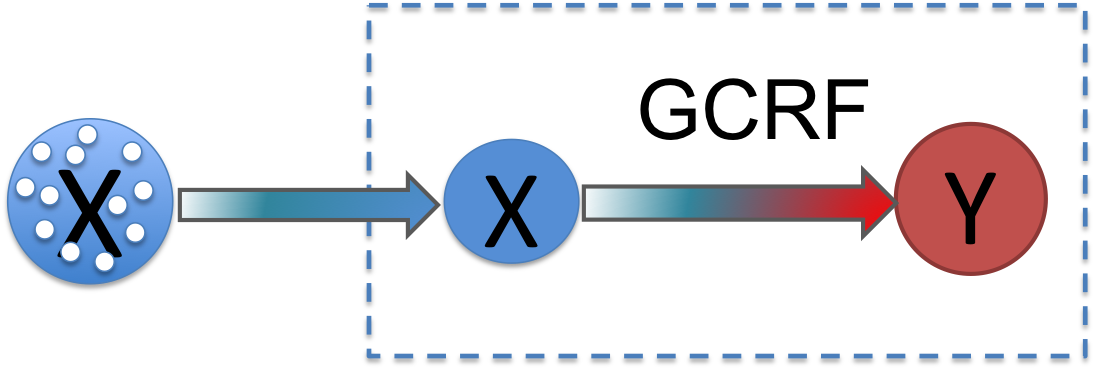}
		\end{center}
		\caption{Using GCRF model on the existing data}
		\label{fig:po_regression}
	\end{subfigure} \\
	\begin{subfigure}{.33\textwidth}
		\begin{center}
			\includegraphics[width=\linewidth]{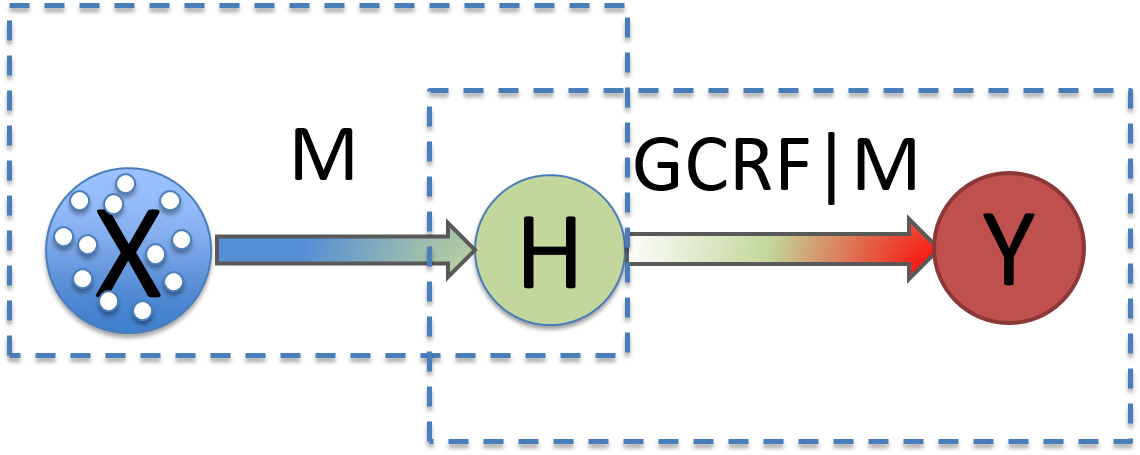}
		\end{center}
		\caption{Applying GCRF on a transformed representation $H$ obtained by unsupervised approach 
			%  unsupervisedly learned input transformation $\mathcal{M}$
		}
		\label{fig:po_unsup_sup}
	\end{subfigure} \\
	\begin{subfigure}{.33\textwidth}
		\begin{center}
			\includegraphics[width=\linewidth]{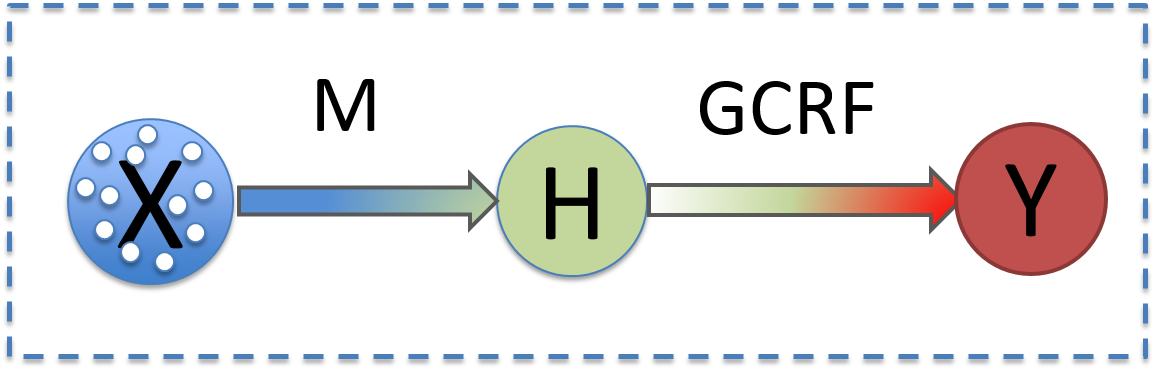}
		\end{center}
		\caption{Joint supervised learning of the input transformation $\mathcal{M}$ and the GCRF model.}
		\label{fig:po_sup_sup}
	\end{subfigure}%
	\caption{Alternatives for applying GCRF on partially observed graphs}
	\label{fig:po_alternatives}
\end{figure}

Given that we are dealing with a regression on data with underlying structures, we are using a structured regression model that has shown recent success in many applications, the GCRF model \cite{Radosavljevic2010,Radosavljevic2014}. Our proposed approach employs learning feature representations to improve predictive power of this method, as well as to handle existing data deficiency. Thus, in terms of robust modeling of explanatory feature mappings and desired predictive task, we could employ several strategies:  
a) Predictive modeling on a complete set (or subset) of \textit{existing raw data}, ignoring partially observed nodes (schematics displayed in Figure~\ref{fig:po_regression}), 
b) \textit{Unsupervised approach}: a common approach where features are learned in an unsupervised fashion prior to learning the predictive model (displayed in Figure~\ref{fig:po_unsup_sup}), 
c) \textit{Supervised approach}: where features are learned simultaneously with the predictive model (displayed in Figure~\ref{fig:po_sup_sup}).
In this section we briefly discuss some of the state-of-the-art approaches related to unsupervised and supervised learning of input feature representations.

\paragraph{Unsupervised feature learning} Many feature learning tasks are defined as unsupervised learning problems. For example, recent success of Deep Restricted Boltzman Machines \cite{tieleman2008} or Deep Autoencoders \cite{hinton2006reducing} has shown benefits of unsupervised feature learning. However, the main limitation of unsupervised feature learning is that it constrains to a parameter space of allowed solutions \cite{bengio2009}, and as such may not necessarily be optimized for the actual problem at hand.
Typically, unsupervised feature learning methods determine the mapping $\mathcal{M}$ (Figure~\ref{fig:po_unsup_sup}) by optimizing an unsupervised objective (e.g., minimizing the reconstruction error), and afterwards a prediction algorithm can be applied to such transformed data. 

Although this transformation $\mathcal{M}$ could capture the underlying structure, it does not necessarily capture the objective of the overall regression (e.g. maximizing the log likelihood of a regressor). Therefore the process involves two objectives: a) minimizing the reconstruction error as an unsupervised objective and b) maximizing the marginal likelihood as a supervised objective. 
Some studies have demonstrated benefits of training CRFs on feature representations learned by unsupervised learners for classification problems \cite{do2010,mohamed2011}.
In this study we will use this type of approach as one of baselines for structured regression and show that our proposed method outperforms such models in terms of accuracy.

\paragraph{Supervised feature learning} In the area of supervised feature learning, several approaches were proposed for the CRF based classification  \cite{djuric2014,do2010,maaten2011,mahajan2006,quattoni2007,Radosavljevic2014}, where benefits were demonstrated on various applications. 
For example, learning of hidden states (or units) between explanatory variables $X$ and response variable $y$ is considered \cite{maaten2011,quattoni2007}, where this model is used for object detection and gesture recognition \cite{quattoni2007} and  for optical character recognition, text classification, protein structure prediction, and part-of-speech tagging \cite{maaten2011}. The approach is also applied to a phone classification task \cite{mahajan2006}, and to ad targeting \cite{djuric2014}. Success on a large variety of tasks has provided enough evidence that such methodology, if applied to continuous CRF's, could improve the model's representational power as well.

A different approach to modeling hidden units is to use neural networks architectures in the association potentials of the CRFs. This approach has shown benefits for both classification \cite{do2010} and regression \cite{Radosavljevic2014}. However, these models were either incapable of modeling complex relationships of response variables (only used a linear chain or a tree structure) or the interaction potential of the CRF's used predefined network structure as input, independently of other explanatory variables. In our approach, we propose using a neural architecture for the supervised mapping, on top of which both representation, as mapping $X \rightarrow y$, and a general graph structure are learned simultaneously.

In our experiments different related published approaches (described in Section~\ref{sec:baseline_models_experimental_setup}) are used as baselines for comparison to the proposed supervised feature learning method. We provide evidence in several case studies for two prediction tasks that a supervised strategy is not only more accurate, but is more robust when applied to partially observed data.

\section{The Model} \label{sec:model}
Here, we first describe the Gaussian Conditional Random Fields (GCRF) model and provide its interpretation in Section~\ref{sec:gcrf_model}. Further, we specify the proposed Deep Feature Learning GCRF (DFL--GCRF) model in Section~\ref{sec:DFLGCRF} and define feature embedding via neural mapping (Section~\ref{sec:mapping}), as the chosen mapping function for deep feature learning model.

\subsection{Gaussian Conditional Random Fields}
\label{sec:gcrf_model}
Gaussian Conditional Random Fields (GCRF) \cite{Radosavljevic2010} is a discriminative structured regression model. The model captures both the network structure of variables of interest ($y$) and relationship between attribute values of the nodes ($X$) and the target variable $y$. It is a model over a general graph structure (not only chains or trees), and can represent the relationships of the nodes as a function of time, space, or any user-defined structure. It models the structured regression problem by estimating a joint continuous distribution over all nodes. GCRF takes the following log-linear form:

\begin{multline*} \label{eq:GCRF_orig}
P(y| X) = \frac{1}{Z(x,\alpha,\beta)}exp(\phi(y,X,\alpha,\beta)) = \\
\frac{1}{Z(x,\alpha,\beta)}exp(\sum_{i=1}^{N}A(\alpha, y_i , X) + \sum_{i \sim j}I(\beta, y_i , y_j , X)) = \\
\frac{1}{Z}exp(-\sum_{i=1}^{N} \sum_{k=1}^{K}\ 
{\alpha _{k}} (y_i - R_k(X))^2 \\
- \sum_{i \sim j} \sum_{l=1}^{
	L}{\beta_l} {S_{ij}}^{(l)}(y_i - y_j)^2) \numberthis
\end{multline*}

The first part of the log-linear form $A(\alpha, y_i , X) = -\sum_{k=1}^{K} \alpha _{k}$  $(y_i - R_k(X))^2$ is called \emph{association potential} and it aims to model associations $X \rightarrow y_i$ using $K$ different functions $R_k(X)$, which we will call unstructured predictors, as they are modeling these associations independently by learning from data or by using domain knowledge. Parameters of the association potential $\alpha_k$ are learned as degrees of belief towards each unstructured regressors. Given by the squared error $\sum_{i=1}^N(y_i - R(X))^2$, 
larger belief $\alpha$ is learned to correspond to the more accurate unstructured predictor.

The second part $I(\beta, y_i , y_j , X) =- \sum_{l=1}^{L}\beta_l {S_{ij}}^{(l)}(y_i - y_j)^2$ is called \emph{interaction potential} and its goal is to utilize 
%structure between labels defined over the similarity matrix $S$. 
a graph structure $S$, that should be a weighted undirected network whose edges $S_{ij}$ denote how similar two nodes are, or more precisely, how similar their response values $y_i$ and $y_j$ are.
%, that defines the positively weighted undirected graph of nodes. 
Parameters $\beta$ are learned as degrees of belief towards similarity metrics and their values are governed by the product of similarity metric and squared distance $\sum_{i \sim j}S_{ij}(y_i - y_j)^2$. 
If this distance is small, relative value of $\beta$ will be larger and the entire model will take the structure as an important source of information.

The normalization term 
\begin{equation}
Z(x,\alpha,\beta) = \int_{y} \exp( \phi (y, X, \alpha, \beta)) dy
\end{equation}
and in general case, estimating this term is intractable. However, using quadratic feature functions, as demonstrated in Eq.~\ref{eq:GCRF_orig}, enables an elegant representation of the log-linear form as a multivariate Gaussian distribution \cite{Radosavljevic2010}:
\begin{equation}
P(y|X) = \frac{1}{(2\pi)^{\frac{N}{2}}\mid \Sigma\mid^\frac{1}{2}}exp\left(-\frac{1}{2}(y - \mu)^{T}\Sigma^{-1}(y - \mu) \right ) ,
\end{equation}
which allows efficient convex optimization.
Here, $\Sigma^{-1}$ represents the diagonally dominant inverse covariance matrix, and for this model takes the form:
\begin{equation}
\Sigma^{-1} = \left\{\begin{matrix}
2\sum_{k=1}^{K}\alpha_{k} + 2\sum_{g}\sum_{l=1}^{L}\beta_l S_{ig}^{(l)}(x) , i=j 
\\
-2\- \sum_{l=1}^{L}\beta_l S_{ij}^{(l)}(x) , i\neq j
\end{matrix}\right.
\end{equation}

The posterior mean is given by $\mu = \Sigma  b,$ where $b$ is defined as 
\begin{equation}
b_i = 2\left(\sum_{k=1}^{K}\alpha_{k} R_{k}(X) \right ).
\end{equation}

\subsubsection{Learning and inference}
The learning task is to optimize parameters $\alpha$ and $\beta$ by maximizing the conditional log--likelihood $ \mathcal{L}$,
\begin{equation}
(\hat{\alpha }, \hat{\beta}) = \underbrace{argmax}_{\alpha, \beta} \mathcal{L}=  \underbrace{argmax}_{\alpha, \beta}logP(y|X;\alpha,\beta).
\end{equation}

Parameters $\alpha$ and $\beta$ are learned by a gradient-based optimization. Gradients of the conditional log-likelihood are:
\begin{multline*} \label{eq:deriv_lik_alpha}
\frac{\partial \mathcal{L} }{\partial \alpha_k} = -\frac{1}{2}(y-\mu)^{T}\frac{\partial \Sigma^{-1} }{\partial \alpha_k}(y-\mu) + \\ ( \frac{\partial b^{T}}{\partial \alpha_k} - \mu^{T}\frac{\partial \Sigma^{-1} }{\partial \alpha_k})(y-\mu) + Tr(\Sigma\frac{\partial \Sigma^{-1} }{\partial \alpha_k}) \numberthis
\end{multline*}
\begin{multline*} \label{eq:deriv_lik_beta}
\\ \frac{\partial \mathcal{L} }{\partial \beta_l} = -\frac{1}{2}(y+\mu)^{T}\frac{\partial \Sigma^{-1} }{\partial \beta_l}(y-\mu) + Tr(\Sigma\frac{\partial \Sigma^{-1} }{\partial \beta_l}) \numberthis
\end{multline*}
Maximizing the conditional log--likelihood is a convex objective, and can be optimized using standard Quasi-Newton optimization techniques. Constraint of positive-semi definiteness of matrix $\Sigma^{-1}$ ensures that the distribution is Gaussian. Therefore, to make the optimization unconstrained, the exponential transformation of parameters $\alpha_{k} = e^{u_{k}}$ and $\beta_{l} = e^{v_{l}}$ is used in GCRF \cite{Radosavljevic2010}.

In this model, prediction is governed by two parts, the association and interaction potentials. The association potential guides the main prediction power of the GCRF model
and clearly, the more accurate the unstructured models are, the more GCRF will assimilate those predictions. On the other hand, as these unstructured predictors usually do not take into account the structure information, interaction potential will compensate that by introducing similarity matrix $S$, and bringing the predictions of the connected nodes closer together. A combination of the two potentials provides better accuracy than the unstructured predictors alone. However, as both unstructured predictor $R$ and similarity matrix $S$ are given (learned prior to GCRF model learning) they introduce a bias in the model.
That is why in this study we propose a more complex, non-convex generalization of the GCRF model where $R$ and $S$ are learned within the GCRF framework. This extension will remove the bias of using pre-trained inputs. However, the bias will still be present in the form of chosen $R$ and $S$ functions. The trade off between model convexity and performance is a well studied topic and a number of studies have pointed out that convexity does not necessarily lead to the more powerful models \cite{bengio2007,lecun1998}. The new model will optimize the $R$ and $S$ for the overall regression goal and as such will improve its representational power.

\subsection{Feature learning with the GCRF model} 
\label{sec:DFLGCRF}

Most existing approaches often rely on a two-step process where a latent representation of explanatory variables is trained first, and its output is used to generate potentials for the structured predictor. This piece-wise training is, however, suboptimal, as the deep features are learned while ignoring the dependencies between the variables of interest. However, when learned jointly they can improve their predictive power by exploiting complementary information to build on the available data, and thus be beneficial for the overall regression task.

In order to implement this approach to the existing GCRF framework \cite{Radosavljevic2010} and show its benefits, we have extended GCRF by:
\begin{itemize}
	\item learning unstructured predictors $R(X, \theta)$ and similarity functions $S(x_i,x_j,\psi)$ together with learning $\alpha$ and $\beta$ parameters of GCRF, rather than using them as pre-trained;
	\item defining a feature mapping function $\mathcal{M}(X, \xi)$ that takes available explanatory variables $x_i \in \rm I\!R^{m}$, for $i = 1,...,N$ and maps them into $\rm I\!R^{h}$. \footnote{The dimension of latent features $h$ is arbitrarily chosen by the user} Both unstructured predictors and similarity metrics will be dependent on newly generated features, and we can formalize them as $R(\mathcal{M}(X, \xi), \theta)$ and $S(\mathcal{M}(X, \xi),\psi)$.
\end{itemize}
As our model performs feature learning together with learning input--output mapping and complex outputs' relations in a deep framework, we refer to this model as Deep Feature Learning GCRF model (DFL--GCRF). The diagram of the DFL--GCRF model is given in Figure~\ref{fig:NNMapping}.

This approach adds an additional three groups of parameters that are trained simultaneously with previously defined parameters $\alpha$ and $\beta$. In order for this extension to work, the unstructured predictor $R(x, \theta)$, similarity function $S(x_i,x_j,\psi)$ and feature mapping function $\mathcal{M}(x_i, \xi)$, need to be differentiable functions w.r.t. their parameters.

The final log-linear form of the DFL--GCRF:
\begin{multline}
\label{eq:po_gcrf}
P(y| X)= \frac{1}{Z}exp(-\sum_{i=1}^{N} \sum_{k=1}^{K}\
\alpha _{k} (y_i - R_k(\mathcal{M}(X,\xi),\theta_{k}))^2 \\
- \sum_{l=1}^{L}\sum_{i \sim j} \beta_l {S_{ij}}^{(l)}(\mathcal{M}(X,\xi), \psi_{l}) (y_i-y_j)^2
\end{multline}

Then the inverse covariance (precision) matrix $\Sigma^{-1}$ changes its form to: 
\begin{equation}\label{eq:po_Q}
\Sigma^{-1} = \left\{\begin{matrix}
2\sum_{k=1}^{K}\alpha_{k} + 2\sum_{g}\sum_{l=1}^{L}\beta_l S_{ig}^{(l)}(\mathcal{M}(X,\xi), \psi_{l}) , i=j
\\
-2\sum_{l=1}^{L}\beta_l S_{ij}^{(l)} (\mathcal{M}(X,\xi), \psi_{l}), i\neq j
\end{matrix}\right.
\end{equation}

as well as $b$:
\begin{equation}\label{eq:po_b}
b_i = 2\left(\sum_{k=1}^{K}\alpha_{k} R_k(\mathcal{M}(X,\xi),\theta_{k}) \right )
\end{equation}
The first moment of the multivariate Gaussian is obtained in the same way as before: $\mu = Q^{-1} b$.

This form of the model has the potential of using any linear or non-linear differentiable unstructured predictor, and any positive differentiable similarity function (the choice of these functions are presented in Section~\ref{sec:experim_setup}). However, joint optimization of the unstructured predictors and similarity metric with the GCRF doesn't allow for a convex optimization objective. An additional layer of complexity is introduced with the mapping function $\mathcal{M}(X,\xi)$. We describe solution for this complex optimization in the following section. With these additions we obtain a highly powerful and robust algorithm for modeling complex relationships.

\subsubsection{Learning and Inference}
The learning task is now to optimize parameters $\alpha$, $\beta$, $\theta$, $\psi$, $\xi$  by maximizing the conditional log-likelihood,
\begin{equation}\label{eq:po_learning}
(\hat{\alpha},\hat{\beta}, \hat{\theta}, \hat{\psi}, \hat{\xi)} = \underbrace{argmax}_{\alpha,\beta, \theta, \psi, \xi}logP(y|X).
\end{equation}
The modeled distribution is a multivariate Gaussian. Therefore, even though the objective function is no longer convex, it is a smooth function. As such, the parameters can still be learned by the gradient based methods with \textit{warm start} techniques to avoid obvious local minimums \cite{bengio2012}. The partial derivatives of the conditional log likelihood w.r.t. parameters $\alpha$ and $\beta$ are given in in the Eq. \ref{eq:deriv_lik_alpha} and Eq. \ref{eq:deriv_lik_beta}, and derivatives w.r.t parameter $\theta_k$ will be:

\begin{equation}
\label{eq:po_partR}
\frac{\partial \log P}{\partial \theta_{k}}=\frac{\partial \log P}{\partial{R_k}} \frac{\partial{R_k}}{\partial \theta_{k}}
\end{equation}
where $\frac{\partial \log P}{\partial{R_k}} = 2\alpha_k^T(y - \mu)$ and the second component depends on the chosen function $R_k$. The derivatives w.r.t parameters $\psi_{l}$ are
\begin{multline}
\label{eq:po_partS}
\frac{\partial \log P}{\partial \psi_{l}}=-\frac{1}{2}(y + \mu)^{T}\frac{\partial \Sigma^{-1}}{\partial{S_l}} \frac{\partial{S_l}}{\partial \psi_{l}}(y - \mu)+\\
\frac{1}{2}Tr({\Sigma\frac{\partial \Sigma^{-1}}{\partial{S_l}} \frac{\partial{S_l}}{\partial \psi_{l}}})
\end{multline}
where derivatives depend on the chosen function $S_l$. Finally, derivatives w.r.t parameters $\xi$ are
\begin{multline}
\label{eq:po_partM}
\frac{\partial \log P}{\partial \xi} = -\frac{1}{2}(y - \mu)^{T}\frac{\partial \Sigma^{-1}}{\partial{\mathcal{M}}} \frac{\partial{\mathcal{M}}}{\partial \xi}(y - \mu)+\\(\frac{\partial b}{\partial{\mathcal{M}}} \frac{\partial{\mathcal{M}}}{\partial \xi}-\mu^{T}\frac{\partial \Sigma^{-1}}{\partial{\mathcal{M}}} \frac{\partial{\mathcal{M}}}{\partial \xi})(y-\mu)+
\frac{1}{2}Tr({\Sigma\frac{\partial \Sigma^{-1}}{\partial{\mathcal{M}}} \frac{\partial{\mathcal{M}}}{\partial \xi}}),
\end{multline} 
where derivatives depend on the chosen input transformation $\mathcal{M}$, which will be discussed in detail in Section~\ref{sec:mapping}.
The procedure for a gradient based optimization of the DFL--GCRF model is provided in the Algorithm~\ref{alg:ne_gcrf}.

\begin{algorithm}
	\caption{DFL--GCRF optimization procedure}
	\label{alg:ne_gcrf}
	\begin{algorithmic}[]
		\State \textbf{Input:} Training data \textbf{X}, \textbf{y}
		\State Initialize $\theta, \psi, \alpha, \beta$, $\xi$
		\begin{enumerate}[label=(\alph*)]
			\item Estimate $\xi$ by an unsupervised feature mapping strategy
			\item Estimate $\theta$ by learning unstructured predictor on mapped input space
			\item Estimate $\psi$ by optimizing similarity for given nodes
			\item Estimate $\alpha, \beta$ by optimizing the GCRF model that uses unstructured predictor and similarity learned in steps 2(b) and 2(c) as inputs using Equations~\ref{eq:deriv_lik_alpha} and~\ref{eq:deriv_lik_beta}\
		\end{enumerate}
		\Repeat
		\State Apply gradient based optimization to estimate all parameters using Equations~\ref{eq:deriv_lik_alpha},~\ref{eq:deriv_lik_beta},~\ref{eq:po_partR},~\ref{eq:po_partS} and~\ref{eq:po_partM}
		\Until{convergence}
	\end{algorithmic}
\end{algorithm} 

To avoid overfitting, which is a common problem for maximum likelihood optimization, we added regularization terms for $\alpha$, $\beta$, $\theta$, $\psi$, $\xi$ to the log-likelihood to penalize large outputs of the parameters. The maximum posterior estimate of $y$ is then obtained by computing the expected value $\mu$:
$ \hat y = \underbrace{argmax}_{y}P(y|X)=\mu $.

In Section~\ref{sec:experim_setup} a particular implementation of the architecture used in our experiments will be described in more details, including the choice of $R$ (unstructured predictors), $S$ (similarity) and $\mathcal{M}$ (mapping) functions.

\subsubsection{Neural Mapping for GCRF}
\label{sec:mapping}
\begin{figure}[h]
	\centering
	\includegraphics[width=0.45\textwidth]{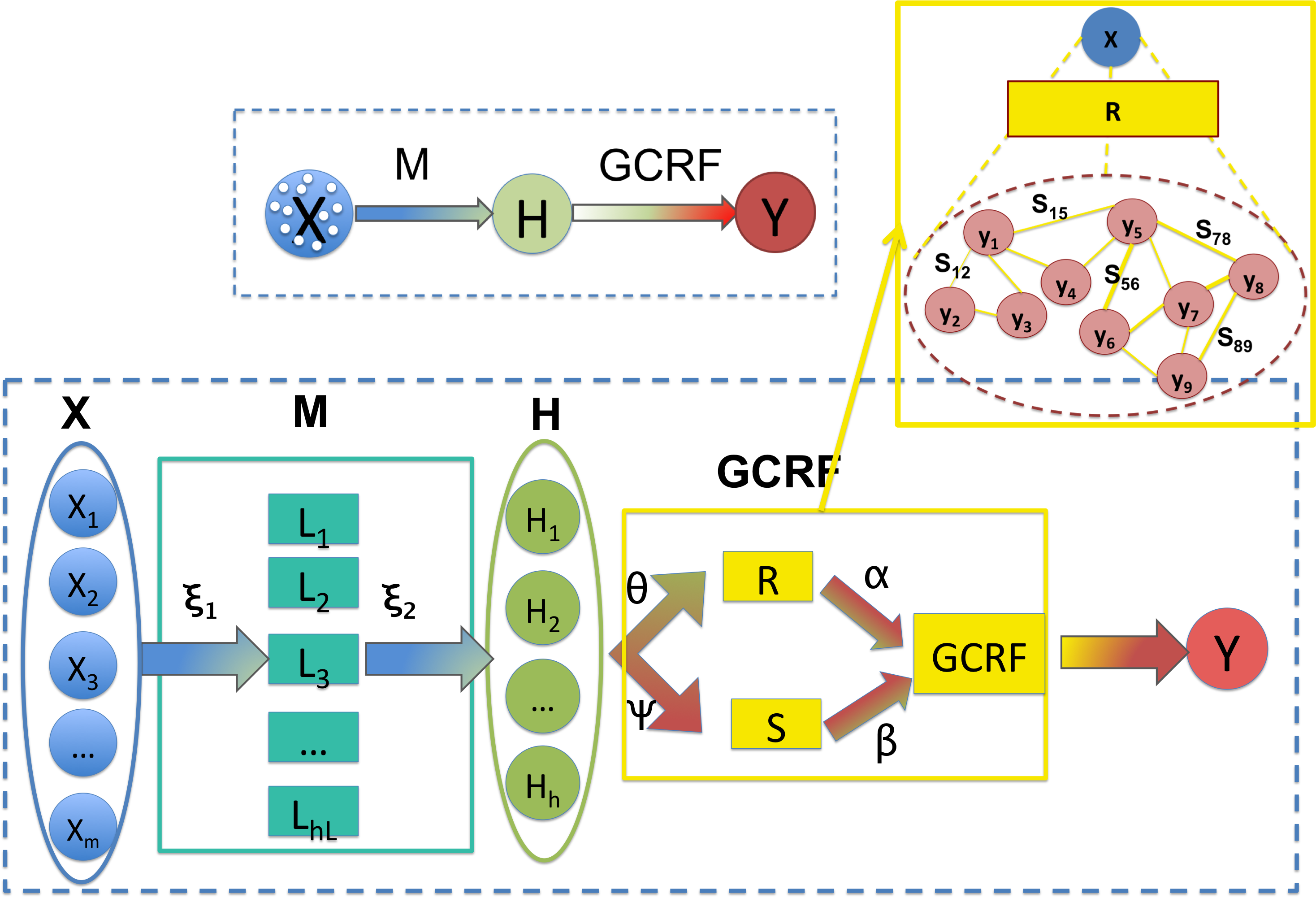}
	\caption{\textbf{Deep Feature Learning GCRF Framework}: Mapping function in our experiments is a neural network. This neural network will map explanatory attributes to attributes of a latent layer ($H$) and on such mapped data, GCRF is applied with linear regression as an unstructured predictor ($R$) and Gaussian kernel as a similarity function ($S$). Parameters of the mapping function ($\xi$), as well as parameters of the unstructured predictor ($\theta$) and the similarity function ($\psi$) are learned together with GCRF objective function and its parameters ($\alpha$ and $\beta$).}
	\label{fig:NNMapping}
\end{figure}
We consider the general setting where $\mathcal{M}(X,\xi)$ can be any arbitrary function of $\xi$ and $X$. 
Options for $\mathcal{M}(X,\xi)$ reported in literature include different matrix factorization approaches on feature matrix $X$ \cite{levy2014,zhou2014} or various kernels \cite{cho2009,qin2014}. 
Matrix factorization approaches as well as different kernel approaches often fail to outperform neural feature mappings on tasks of feature learning \cite{perozzi2014deepwalk}. 
Other approaches include dictionary learning \cite{mairal2009online}, which often includes $L_1$ regularization to enforce sparsity, and as a consequence affects smoothness of the optimization function \cite{wytock2013}. 
These methods require slow, dedicated methods for optimization, thus limiting their applicability to smaller datasets.
In this study, a neural feature embedding architecture (Figure~\ref{fig:NNMapping}) is used, as recent advances in the field of feature learning show promising results when applying such embeddings to a variety of tasks in many domains \cite{bengio2003,do2010,mikolov2013}. 
Mapping function $\mathcal{M}(X,\xi) = \sigma(\xi X + b)$, with Sigmoid $\sigma(g) = \frac{1}{1+e^{-g}}$ and a matrix of mapping weights $\xi$, constitutes the first layer of deep architecture of DFL--GCRF framework (implementation details are given in Section~\ref{sec:experim_setup}). 

Our hypothesis for such defined framework is that examples with partially missing inputs yield nearly equally good hidden representation as completely observed inputs.
Previously, neural mapping approaches were successfully applied to the task of reconstructing corrupted input features, and theoretical analysis from several perspectives was provided on its validity \cite{vincent2008extracting}. The difference comparing to this approach is that we expect the neural mapping $\mathcal{M}(X,\xi)$ to learn response variable values $y$ from deficient explanatory variables $X$ supervisedly, instead of reconstructing the explanatory variables $X$, first, in an unsupervised manner. Experimental results showed that the proposed model successfully outperformed this baseline model and thus, backed up our hypothesis. 
Finally, our approach can be formalized as learning a stochastic mapper which transforms mixture of full and deficient data points to a manifold that allows for a linear mapping to the response variable $p(y|(H|(X_d,X_p)))$, with $H$ being a representation of heterogeneous inputs that captures the main variations in the data w.r.t. response variable. 

\section{Experimental setup}
\label{sec:experiments}
In this section an experimental setup for the proposed and baseline methods are described.  

\subsection{The proposed method setup}
\label{sec:experim_setup}
In our experiments the DLR-GCRF uses a neural mapping as shown in Figure~\ref{fig:NNMapping} with the following architecture: an input layer of dimensionality $\rm I\!R^{m}$, one hidden layer of dimension $\rm I\!R^{h_{L}}$, and $\rm I\!R^{h}$ dimensional output with distributed features. The number of neurons in a hidden layer of this neural mapping is $h_{L}\approx \frac{N}{\gamma * (m+ h)}$ \cite{hagan1997neural}, where the number of outputs of this neural mapping is $h$ and it is $3$ in our experiments, and $\gamma$ is chosen arbitrary in  $[2-10]$ range. 

The choice of unstructured predictor $R$ is a linear model, which uses the first two learned distributed features, while the choice of $S$ is a Gaussian kernel learned on the third distributed feature.

\subsection{Baseline models setup}
\label{sec:baseline_models_experimental_setup}
To evaluate the effectiveness of the DFL--GCRF  model, we are comparing it to ten alternative methods from groups of models that are using the complete set or a subset of existing data as inputs and models that use latent features learned in an unsupervised manner.

First, we test the performance of the following  baseline methods that learn their parameters only on the observed part of the data (ignoring the cases with missing inputs), as described in the Figure~\ref{fig:po_regression}.
\begin{itemize}
	\item Linear Regression (iLR): We applied the unstructured linear predictor which captures the linear influence of explanatory variables $X$ to a response variable $y$;
	\item Gaussian Processes Regression (iGP) \cite{rasmussen2006gaussian}: We tested the GP model with a Gaussian Kernel $GK(x_i,x_j)$. Kernel optimized via GP objective function was further used as a network structure for structured models;
	\item Gaussian Conditional Random Fields (iGCRF): We also evaluate the GCRF model which utilizes the unstructured predictor and the available structure (in our experiments structure is node covariates learned with Gaussian Kernel). %We refer to this model as the i-GCRF model. 
\end{itemize}

Further, we compared the proposed model  to several models from the group of unsupervised feature learning methods, as shown in Figure~\ref{fig:po_unsup_sup}: 
here, we apply one of the mapping functions and afterwards learn the GCRF regression model on such a mapped dataset ($H$). 
To isolate all other effects, we always used the same set-up for the structured GCRF model. This consists of an unstructured predictor of the GCRF model learned on mapped feature space using a linear model and interactions modeled via a Gaussian kernel function. Baseline mapping functions in this category that we applied are:
\begin{itemize}
	\item Deep Autoencoders (DAE) \cite{hinton2006reducing, vincent2008extracting}: DAE aims to automatically learn features from unlabeled data by minimizing the input reconstruction error, namely, by learning a compressed, distributed representation (encoding) for a set of input data, typically for the purpose of dimensionality reduction;
	\item Principal Component Analysis (PCA) \cite{van2009dimensionality}: PCA aims to find a linear projection of high dimensional data into a lower dimensional subspace such that the variance is retained and the least square reconstruction error is maximized;
	\item Neural Mapping (NM) is learned in a supervised manner by optimizing a neural network (NN) for regression -- mapping is defined as the last hidden layer of the neural network. The architecture of the NM is exactly the same as that of the neural mapping in the DFL--GCRF model. Note that this mapping is learned with the neural network optimization function and not with the GCRF optimization function;
	\item Zero imputation: In the situation when data are missing, a $0$ value is imputed. As baselines in this category we used  LR--0, GP--0 and GCRF on such 0--based imputed dataset.
\end{itemize}

\noindent The effectiveness of the proposed (DFL--GCRF) vs baseline methods (NM + GCRF, NN, DAE + GCRF, PCA + GCRF, GCRF, LR--0, GP--0, iGCRF, iLR and iGP) is evaluated on two applications described in Section~\ref{sec:data} and the metric used for evaluation is the coefficient of determination  defined as
$
R^{2} = 1 - \frac{\sum_{i} (y_{i} - \mu_{i})^{2}}{\sum_{i} (y_{i} - \hat{y})^{2}},
$
where $y_i$ and $\mu_i$ are true and predicted value for customer $c_i$, and $\hat{y}$ is the mean value for all customers in $\mathcal{C}$. 
We limit the values of $R^2$ to $[0,1]$ scale, as we treat predictors with negative $R^2$ performance as useless, while predictors that obtain $R^2 = 1$ are considered to be a perfect fit to the data.

\section{Experimental results}
\label{sec:experimental_evaluation}
In this section the results are shown for: (1) predicting the customers' visit frequency in the following month, and (2) predicting individual customer's ticket in the upcoming quarter using their partially available demographic data, as well as their purchase history. 
%The proposed method has been evaluated on datasets from other domains, as well, showing similar results as those presented in this section, however due to the space limitation, we are omitting them from this paper. 

\subsection{Prediction of visit frequency}
\label{sec:latitude}
The first company bases its membership on a monthly fee such that customers can use a certain number of provided services (depending on the program they signed up for). To provide ''one-to-one'' type messaging and added value that is unique to the customer, the aim is to estimate how often each customer uses purchased services in the following month. Even though these services are free of charge at the visit (for customers who paid the monthly fee), they often spend money on side products and services during the visit and bring additional revenue to the company. Therefore, with the knowledge of estimates of forthcoming visit frequencies, the company may build additional special offers to incentivize rare visitors or may reward the most loyal customers by, for example, providing instant benefits for a specific upcoming event. Additionally, the company may use this information to further adapt existing programs or educate and remind customers via targeted e-mail campaigns. To evaluate performance of the proposed model versus the alternatives, we conducted a variety of experiments corresponding to several real life situations that might occur with the loyalty program data.
\subsubsection{Predicting visits frequency on fully observed data} \label{sec:latitude_full}
\noindent In the first experiment, we assumed that all demographics prompted by the loyalty program about the company's customers are known. Such data is used to experimentally compare the proposed model with the baseline algorithms described in Section~\ref{sec:baseline_models_experimental_setup}.
%\begin{wraptable}{r}{0.4\linewidth}
	\begin{table}[h!]
	\centering
	\caption{
		Accuracy comparison of DFL--GCRF vs 7 alternatives on complete data for prediction of a customer's visit frequency for the following month. 
	}
	\vspace{5pt}
	{\footnotesize
		\begin{tabular}{cc}
			\toprule
			$model$ & $R^2$ \\
			\midrule
			\rowcolor{lightgray}
			DFL--GCRF & \textbf{0.9147}\\
			NM+GCRF & 0.8793 \\
			\rowcolor{lightgray}
			GCRF & 0.8652 \\
			GP & 0.8582 \\
			\rowcolor{lightgray}
			NN & 0.8525\\
			LR & 0.8502 \\
			\rowcolor{lightgray}
			PCA+GCRF & 0.8350 \\
			DAE+GCRF & 0.8063\\
			
			\bottomrule
		\end{tabular}%
	%	\vspace{55pt}
		\label{tab:latitude_improv_0}%
	}
	\end{table} 
%\end{wraptable}
Results of this experiment are shown in Table~\ref{tab:latitude_improv_0} in terms of $R^2$. 
%(defined in \ref{sec:results_evaluation}) of the DFL--GCRF model compared to alternatives. 
Note that zero imputation and ignoring the missing cases are equivalent when there are no missing data.
\begin{figure*}[htb!]%{0.35\textwidth} %[11]
	\centering
	\begin{subfigure}{.32\textwidth}
		\begin{center}
			\includegraphics[width=\linewidth]{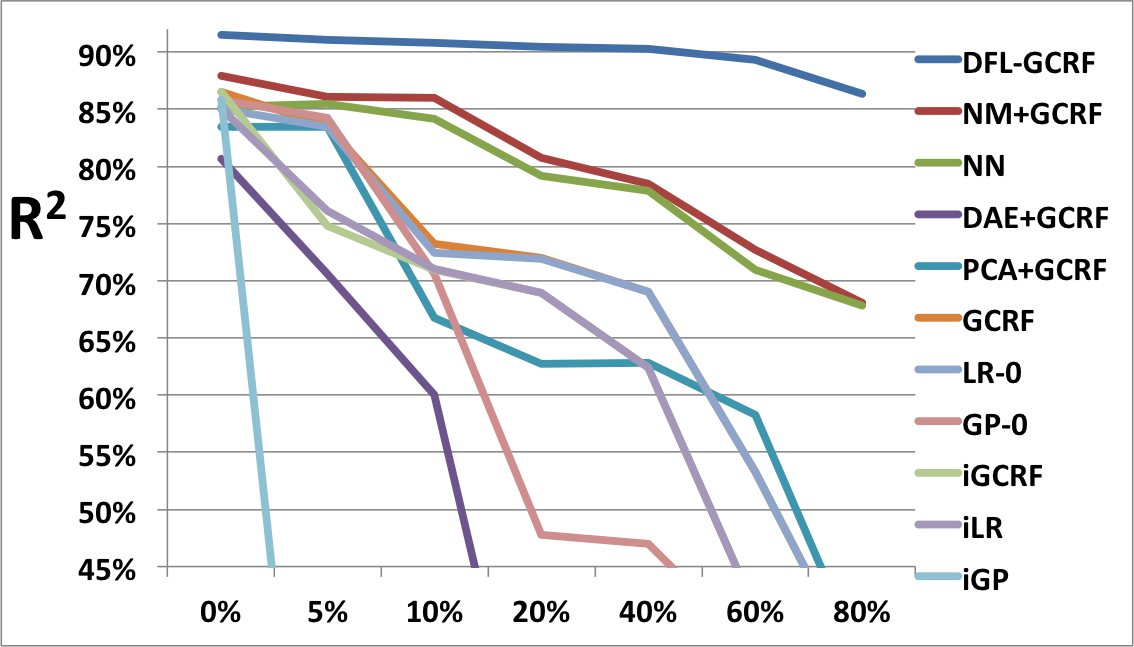}
		\end{center}
		\caption{missing at random}
		\label{fig:latitude_mcar}
	\end{subfigure}
	\begin{subfigure}{.32\textwidth}
		\begin{center}
			\includegraphics[width=\linewidth]{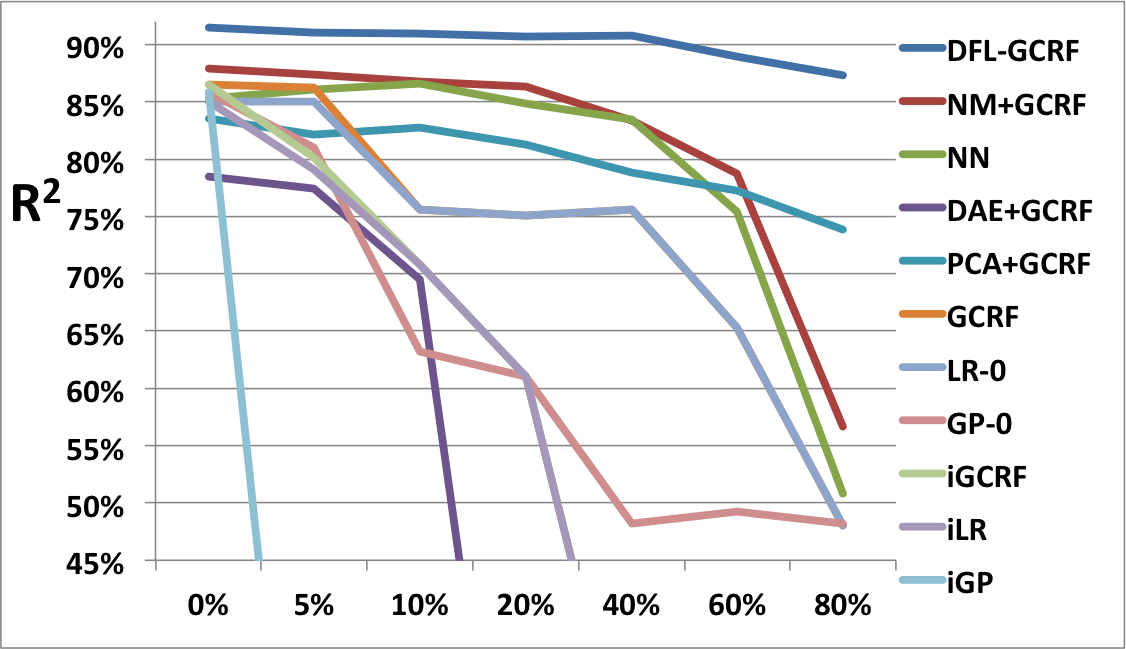}
		\end{center}
		\caption{missing for rare visitors}
		\label{fig:latitude_malo}
	\end{subfigure}
	\begin{subfigure}{.32\textwidth}
		\begin{center}
			\includegraphics[width=\linewidth]{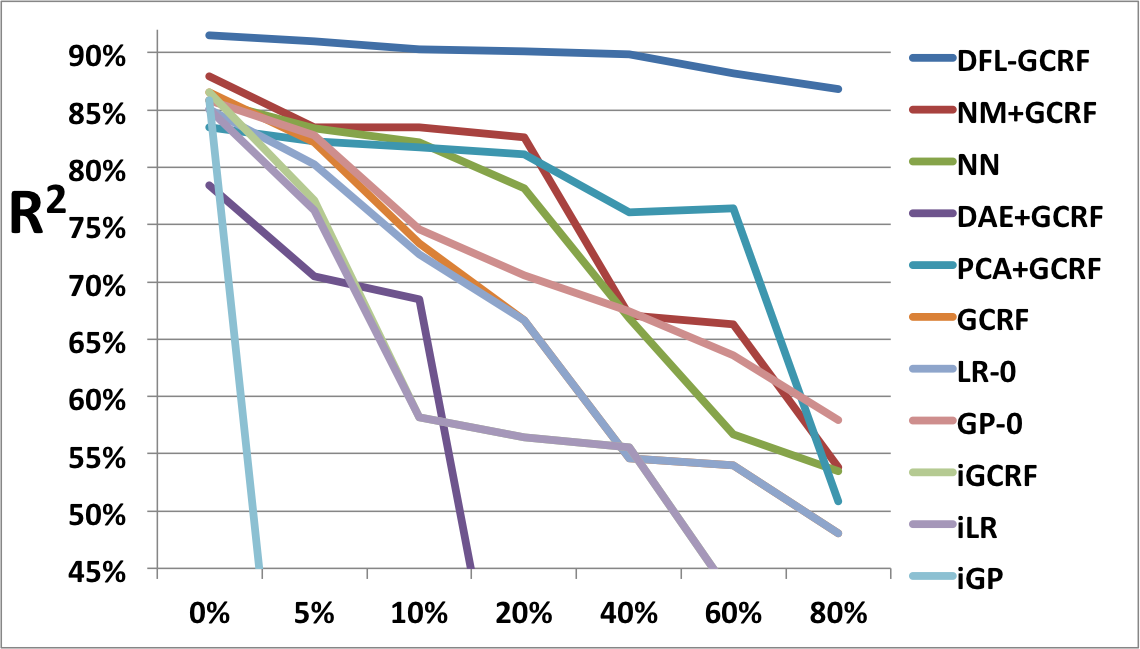}
		\end{center}
		\caption{missing for frequent visitors}
		\label{fig:latitude_puno}
	\end{subfigure}%
	\caption{$R^2$ of predicting visits frequency by DFL--GCRF model vs ten alternatives for up to $80\%$ of demographic information missing by 3 mechanisms.
		%$R^2$ results of the proposed model and benchmarks as we induce up to 40\% of missingness in demographic information by different mechanisms.
	}  
	\label{fig:latitude_missing}
\end{figure*}

From Table~\ref{tab:latitude_improv_0} we observe that the proposed DFL--GCRF model  predicted frequency of visits with more accuracy than any of the alternatives considered. The improvements range from about $4\%$ to $16.5\%$, where every percent of improvement can make a huge difference in terms of decision making when it comes to detecting the users from which the most revenue is generated. Additionally, we can see from the percentages of improvement that GCRF and GP models that account for correlations among the members were more accurate than unstructured predictors (NN and LR). Therefore, modeling relations between customers seems to be beneficial for this prediction task. Our experiments also suggest that unsupervised feature learning models tend to significantly lack accuracy since they are working with lower dimensional inputs. However, methods with neural feature mapping were more accurate. For example, supervised DFL--GCRF and unsupervised feature learning approach NM+GCRF were the best performing models (we also observe that structured model, NM+GCRF, brings improvements to the unstructured NN model). This, henceforth, confirms our assumption of superiority of neural mappings over alternatives as discussed in Section~\ref{sec:mapping}

\subsubsection{Influence of data missingness mechanism on predicting visits frequency} \label{sec:latitude_missing}
In the initial experiments we assumed that all customers were willing to share their demographic data whilst applying for a loyalty program. However, in practice, demographics are lacking in many cases. This is the why we conducted experiments where demographic data is reduced to a fraction of customers. Three types of missingness of demographic data were considered: a) removing demographics of random customers, b) removing data of customers who are least frequent visitors, and c) removing demographic data of the most frequent visitors. These three scenarios were considered for different fractions of missingness ($5\%$ -- $80\%$) in order to characterize their robustness in various situations. The results for all missingness levels are shown in terms of $R^2$ in Figure~\ref{fig:latitude_missing}.

The proposed DFL--GCRF model has outperformed the alternatives and has demonstrated the largest robustness in all three experiments. The overall accuracy improvement was about $5\%$ to $55\%$ vs. nontrivial alternatives for $10\%$ of missing data and about $50\%$ to $368\%$ for $80\%$ of missing data. Some of the baseline models (e.g. iGP) were not better than a mean predictor and are rendered as useless. 
\paragraph{Demographic data missing at random (MAR)}
In the experiment reported at Figure~\ref{fig:latitude_mcar} we examined the situation in which random customers do not reveal their demographics. This way of inducing missingness does not mimic a real process. However, it is an unbiased way of examining the power of the models to handle missing demographic data. The most robust results were obtained by the proposed DFL--GCRF where rather stable accuracy is obtained up to $60\%$ of missing  data. NN and NM+GCRF were also somewhat robust but less accurate than DFL--GCRF. The accuracy in other baseline models considered dropped quite fast (after $10\%$ of missing demographic data), with the exceptions of LR-0 and GCRF (which uses LR-0 as an unstructured predictor) that managed to maintain larger $R^2$ up to $60\%$ of missing demographic data. The unsupervised feature learning models failed even after a few percentages of missing demographic data was induced. The unsupervised DAE approach of reconstructing inputs \cite{vincent2008extracting} under-performed on this task, as shown in Figure \ref{fig:latitude_missing}, and we see that the approach of supervised learning of the mapping function yielded vastly better results, and  thus justified our original hypothesis.
Also, we can see that imputation is a better strategy than ignoring data for each model where we employed these two strategies (LR, GP and GCRF).

\paragraph{Demographic data missing for the least frequent customers}
Experiments reported at Figure~\ref{fig:latitude_malo} examine accuracy when demographic data was missing for rare visitors, which is a common scenario in practice. The customers that are not well engaged with the brand may not be willing to spend their time and energy in filling out forms for registration purposes. The results of this experiment were similar to MAR results. This is due to the fact that majority of the customers we are modeling are actually customers with low frequencies of visits. The main difference between these two results is that when demographic data is missing for the least frequent users the accuracy of the models tend to drop more slowly than in MAR's case (accuracy remains relatively high up to $40\%$ of missingness, rather than up to $10\%$ in MAR). The top three models in these experiments were still neural feature learning models; with the addition of PCA as an unsupervised approach providing good results in the overall missingness induction process. We conclude that the most frequent users contribute the largest amount of variability in the data and thus are the ones from which PCA linear feature mapping can benefit the most.
\paragraph{Demographic data missing for the most frequent customers}
The results shown at Figure~\ref{fig:latitude_puno} are obtained for missing demographic data in a fraction of customers that are frequent visitors (the most loyal ones). We found that this hurts accuracy the most. The main difference versus results shown at Figures~\ref{fig:latitude_mcar} and~\ref{fig:latitude_malo} is a drop in accuracy that occurred as soon as $5\%$ of demographic data for the most frequent users was missing, which is about twice as large, compared to other missingness mechanisms. Additionally, missing values for the most frequent visitors reduced accuracy the most for all of the examined fractions.
\subsection{Prediction of a customer's ticket}

There are several ways in which this company can reward (and therefore incentivize) return customers, which include providing free platinum reward memberships to individuals that spend more than a certain amount over the course of a fiscal quarter, as well as sign--up, preferential, and birthday--related offers and rewards. 
However, while the company encourages customers to disclose birthday information by offering discounts during their birth month, many members still choose to keep this information private. This unwillingness for sharing makes determining future expenditures more difficult, but the company is still interested in selling more to each particular customer, either small or big spender, so it is important to be able to identify them. Our approach allows for accurate prediction of a customer's ticket even when a large fraction of customer demographic information is missing. 
\begin{figure*}[t!]%{0.35\textwidth} %[11]
	\centering
	\begin{subfigure}{.32\textwidth}
		\begin{center}
			\includegraphics[width=\linewidth]{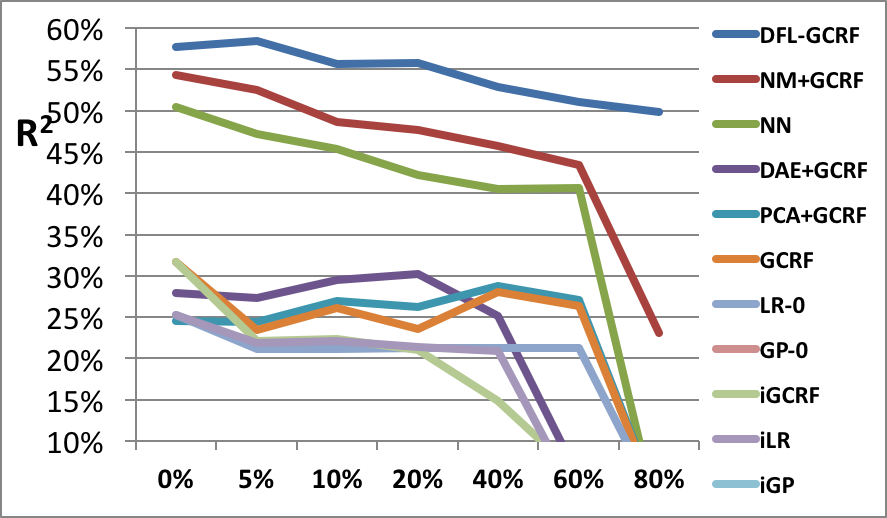}
		\end{center}
		\caption{missing at random}
		\label{fig:lat_mcar}
	\end{subfigure}
	\begin{subfigure}{.32\textwidth}
		\begin{center}
			\includegraphics[width=\linewidth]{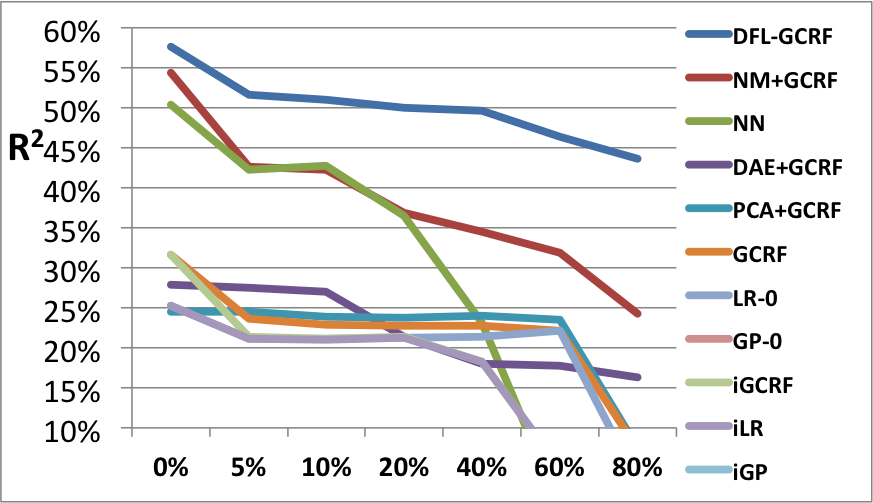}
		\end{center}
		\caption{missing for small spenders}
		\label{fig:lat_smallest}
	\end{subfigure}
	\begin{subfigure}{.32\textwidth}
		\begin{center}
			\includegraphics[width=\linewidth]{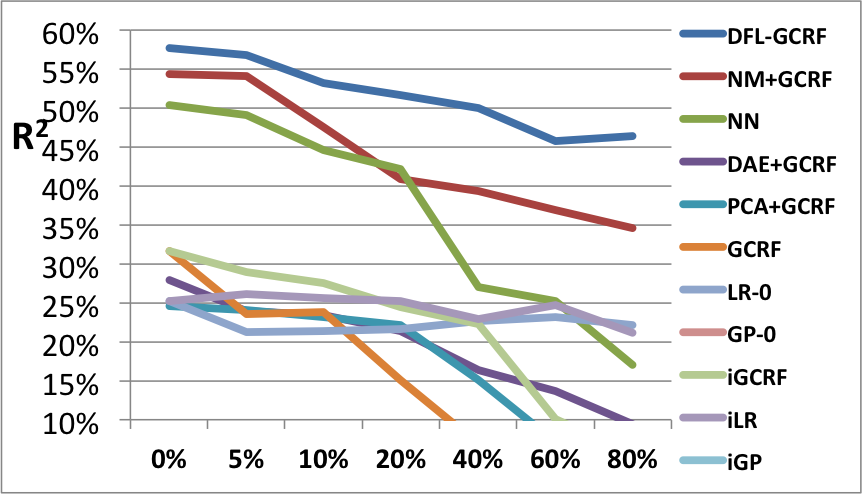}
		\end{center}
		\caption{missing for big spenders}
		\label{fig:lat_largest}
	\end{subfigure}%
	\caption{$R^2$ of predicting customer's ticket by DFL--GCRF model vs ten alternatives for up to $80\%$ of demographic information missing by 3 mechanisms.}
%	\vspace{-10pt}
	\label{fig:ct_missingness}
\end{figure*} 
To evaluate the power of the proposed method versus alternatives for the regression task of predicting customer's ticket for the following quarter, we conducted experiments based on several real life situations that might occur with the loyalty program:
\begin{itemize}
	\item Experiment 1: both demographic data (queried at time of enrollment) and purchase history data is available for all customers
	\item Experiment 2: a random fraction of customers do not provide their demographic data, but their full purchase history is available
	\item Experiment 3: small spenders do not reveal their demographics, but purchasing details are available
	\item Experiment 4: big spenders do not reveal demographics, but purchasing details are available
\end{itemize}

\subsubsection{Predicting customer's ticket on fully observed data} \label{sec:ct_full}
The results for Experiment 1 are summarized in Table~\ref{tab:CT_improv_0}. We observe that the proposed DFL--GCRF model provides a wide range of improvements over alternatives (6\% to 135\%). In practice even a small percent of improvement vastly improves the quality of decision making in this application. We also see that NM+GCRF proves to be the best runner-up model, while GP, which fared well in previous experiments, underperformed on this predictive task (it was worse than a trivial mean predictor). Additionally, the rest of the baselines compare rather unfavorably to the proposed DFL--GCRF, which suggests that yet again, unsupervised feature learning is not an optimal strategy for predictive purposes.

%\begin{wraptable}{l}{0.4\linewidth}
	\begin{table}[h!]
	\centering
	{\footnotesize
		\caption{Accuracy comparison of DFL--GCRF vs 7 alternatives on complete data for prediction of a customer's ticket for the following quarter. }
			\vspace{10pt}
		\begin{tabular}{cc}
			\toprule
			$model$ & $R^2$ \\
			\midrule
			\rowcolor{lightgray}
			DFL--GCRF & \textbf{0.5771} \\
			NM+GCRF & 0.5436\\
			\rowcolor{lightgray}
			NN    & 0.5041 \\
			GCRF  & 0.3165 \\
			\rowcolor{lightgray}
			DAE+GCRF & 0.2789 \\
			LR  & 0.2527 \\
			\rowcolor{lightgray}
			PCA+GCRF & 0.2454\\
			GP  & 0 \\
			\bottomrule
		\end{tabular}%
	%	\vspace{-5pt}
		\label{tab:CT_improv_0}%
	}
	\end{table}%
	
%\end{wraptable}   
The GCRF model that is using LR as an unstructured predictor performs marginally better than unsupervised feature mappings, but it was less accurate than non-linear models.

%\vfill\eject
\vspace{8pt}
\subsubsection{Influence of data missingness mechanisms on customer's ticket estimation}
In order to examine the robustness of DFL--GCRF for the customer's ticket prediction problem we induce up to 80\% of missing values in demographic variables via three mechanisms. The $R^2$ results of ten baseline models and the DFL--GCRF are shown in Figure~\ref{fig:ct_missingness} for these missingess mechanisms.

We observe that, in all three experiments, the overall accuracy of the three neural-based models is the highest, and the gap between those models and the remaining baseline models is much larger for this dataset.
From the non-neural based baselines in all three experiments we see that they are almost unaffected by the increasing missingness in the customer demographic data. We can thus conclude that for this application, these 8 alternative methods failed to utilize demographic information. For three neural-based models, we see that the accuracy drops much faster as compared to experiments reported in Section~\ref{sec:latitude}, even for DFL--GCRF (even though the drop of DFL--GCRF is the smallest compared to the alternative models). In the comparative test of $R^2$ and robustness, DFL--GCRF once again offers the best performance among the models used for different missingness mechanisms and different amounts of missing data, as shown in  Figure~\ref{fig:ct_missingness}.

The improvements of our model as compared to alternatives span an even larger range, starting from 11.85\% and reaching into the thousands in some cases.
\section{Conclusion}
\label{conclusion}

In this paper we introduced Deep Feature Learning GCRF, a powerful deep model for structured regression that learns hidden feature representation jointly with learning complex interactions of nodes in a graph. We have applied this method to two real-world customer engagement problems and provided evidence that the proposed method is capable of learning meaningful features for the purpose of regression, and outperforming other published alternatives developed with a similar aim. Additionally, we have tested the robustness of our method and other baselines when up to 80\% of demographic data is missing by three mechanisms, and thus examined potential cases of missingness that might occur in the actual databases of companies.
In future work we aim to further examine different feature learning approaches aimed to further improve both accuracy and robustness. We additionally aim to extend this approach to detect different groups of similar nodes in a network such that the model would work equally well in highly heterogeneous applications.

\section{Acknowledgment}

This research was supported in part by DARPA Grant FA9550--12--1--0406 negotiated by AFOSR, National Science Foundation through major research instrumentation grant number CNS--09--58854.
The authors gratefully acknowledge use of the data, services and facilities of the Clutch Holdings LLC, Ambler, PA.

%\section{Acknowledgment}
%This research was supported in part by DARPA Grant FA9550-12-1-0406 negotiated by AFOSR.
%The authors gratefully acknowledge use of the data, services and facilities of the Clutch Holdings LLC, Ambler, PA.
\vspace{5pt}
\balance
\bibliographystyle{abbrv}
\bibliography{sigproc}

\begin{thebibliography}{10}

\bibitem{bengio2009}
Y.~Bengio.
\newblock Learning deep architectures for ai.
\newblock {\em Foundations and trends{\textregistered} in Machine Learning},
  2(1):1--127, 2009.

\bibitem{bengio2012}
Y.~Bengio.
\newblock Practical recommendations for gradient-based training of deep
  architectures.
\newblock In {\em Neural Networks: Tricks of the Trade}, pages 437--478.
  Springer, 2012.

\bibitem{bengio2003}
Y.~Bengio, R.~Ducharme, P.~Vincent, and C.~Janvin.
\newblock A neural probabilistic language model.
\newblock {\em The Journal of Machine Learning Research}, 3:1137--1155, 2003.

\bibitem{bengio2007}
Y.~Bengio, Y.~LeCun, et~al.
\newblock Scaling learning algorithms towards ai.
\newblock {\em Large-scale kernel machines}, 34(5), 2007.

\bibitem{cho2009}
Y.~Cho and L.~K. Saul.
\newblock Kernel methods for deep learning.
\newblock In {\em Advances in Neural Information Processing Systems}, 2009.

\bibitem{djuric2014}
N.~Djuric, V.~Radosavljevic, M.~Grbovic, and N.~Bhamidipati.
\newblock Hidden conditional random fields with distributed user embeddings for
  ad targeting.
\newblock In {\em IEEE International Conference on Data Mining}, 2014.

\bibitem{do2010}
T.-M.-T. Do and T.~Artieres.
\newblock Neural conditional random fields.
\newblock In {\em International Conference on Artificial Intelligence and
  Statistics}, pages 177--184, 2010.

\bibitem{Kezunovic2016}
T.~Dokic, P.~Dehghanian, P.-C. Chen, M.~Kezunovic, Z.~Medina-Cetina,
  J.~Stojanovic, and Z.~Obradovic.
\newblock Risk assesment of a transmission line insulation breakdown due to
  lightning and sever weather.
\newblock In {\em HICSS-49}, 2016.

\bibitem{dowling1997customer}
G.~R. Dowling and M.~Uncles.
\newblock Do customer loyalty programs really work?
\newblock {\em Research Brief}, 1, 1997.

\bibitem{Gligorijevic2015}
D.~Gligorijevic, J.~Stojanovic, and Z.~Obradovic.
\newblock Improving confidence while predicting trends in temporal disease
  networks.
\newblock In {\em 4th Workshop on DMMH, 2015 SIAM International Conference on
  Data Mining}, 2015.

\bibitem{Gligorijevic2016}
D.~Gligorijevic, J.~Stojanovic, and Z.~Obradovic.
\newblock Uncertainty propagation in long-term structured regression on
  evolving networks.
\newblock In {\em Thirtieth AAAI Conference on Artificial Intelligence
  (AAAI-16)}, 2016.

\bibitem{hagan1997neural}
M.~Hagan, H.~B. Demuth, M.~Beale, and O.~De~Jesus.
\newblock {\em Neural network design}.
\newblock Martin Hagan, 2014.

\bibitem{hinton2006reducing}
G.~E. Hinton and R.~R. Salakhutdinov.
\newblock Reducing the dimensionality of data with neural networks.
\newblock {\em Science}, 313(5786):504--507, 2006.

\bibitem{lecun1998}
Y.~LeCun, L.~Bottou, Y.~Bengio, and P.~Haffner.
\newblock Gradient-based learning applied to document recognition.
\newblock {\em Proceedings of the IEEE}, 86(11):2278--2324, 1998.

\bibitem{levy2014}
O.~Levy and Y.~Goldberg.
\newblock Neural word embedding as implicit matrix factorization.
\newblock In {\em Advances in Neural Information Processing Systems}, pages
  2177--2185, 2014.

\bibitem{maaten2011}
L.~Maaten, M.~Welling, and L.~K. Saul.
\newblock Hidden-unit conditional random fields.
\newblock In {\em International Conference on Artificial Intelligence and
  Statistics}, pages 479--488, 2011.

\bibitem{mahajan2006}
M.~Mahajan, A.~Gunawardana, and A.~Acero.
\newblock Training algorithms for hidden conditional random fields.
\newblock In {\em 2006 IEEE International Conference on Acoustics, Speech and
  Signal Processing}. IEEE, 2006.

\bibitem{mairal2009online}
J.~Mairal, F.~Bach, J.~Ponce, and G.~Sapiro.
\newblock Online dictionary learning for sparse coding.
\newblock In {\em 26th Annual International Conference on Machine Learning}.
  ACM, 2009.

\bibitem{mikolov2013}
T.~Mikolov, I.~Sutskever, K.~Chen, G.~S. Corrado, and J.~Dean.
\newblock Distributed representations of words and phrases and their
  compositionality.
\newblock In {\em Advances in Neural Information Processing Systems}, pages
  3111--3119, 2013.

\bibitem{mohamed2011}
A.-r. Mohamed, T.~N. Sainath, G.~Dahl, B.~Ramabhadran, G.~E. Hinton,
  M.~Picheny, et~al.
\newblock Deep belief networks using discriminative features for phone
  recognition.
\newblock In {\em 2011 IEEE International Conference on Acoustics, Speech and
  Signal Processing}, pages 5060--5063. IEEE, 2011.

\bibitem{perozzi2014deepwalk}
B.~Perozzi, R.~Al-Rfou, and S.~Skiena.
\newblock Deepwalk: Online learning of social representations.
\newblock In {\em 20th ACM SIGKDD International Conference on Knowledge
  Discovery and Data Mining}, pages 701--710. ACM, 2014.

\bibitem{qin2014}
D.~Qin, X.~Chen, M.~Guillaumin, and L.~V. Gool.
\newblock Quantized kernel learning for feature matching.
\newblock In {\em Advances in Neural Information Processing Systems}, pages
  172--180, 2014.

\bibitem{quattoni2007}
A.~Quattoni, S.~Wang, L.-P. Morency, M.~Collins, and T.~Darrell.
\newblock Hidden conditional random fields.
\newblock {\em IEEE Trans. on Pattern Analysis \& Machine Intelligence}, 2007.

\bibitem{Radosavljevic2010}
V.~Radosavljevic, S.~Vucetic, and Z.~Obradovic.
\newblock Continuous conditional random fields for regression in remote
  sensing.
\newblock In {\em 19th European Conf. on Artificial Intelligence}, 2010.

\bibitem{Radosavljevic2014}
V.~Radosavljevic, S.~Vucetic, and Z.~Obradovic.
\newblock Neural gaussian conditional random fields.
\newblock In {\em Proc. European Conference on Machine Learning and Principles
  and Practice of Knowledge Discovery in Databases}, 2014.

\bibitem{rasmussen2006gaussian}
C.~E. Rasmussen.
\newblock {\em Gaussian processes for machine learning}.
\newblock Citeseer, 2006.

\bibitem{stauss2005customer}
B.~Stauss, M.~Schmidt, and A.~Schoeler.
\newblock Customer frustration in loyalty programs.
\newblock {\em International Journal of Service Industry Management},
  16(3):229--252, 2005.

\bibitem{Stojanovic2015}
J.~Stojanovic, M.~Jovanovic, D.~Gligorijevic, and Z.~Obradovic.
\newblock Semi-supervised learning for structured regression on partially
  observed attributed graphs.
\newblock In {\em SIAM International Conference on Data Mining}, 2015.

\bibitem{tieleman2008}
T.~Tieleman.
\newblock Training restricted boltzmann machines using approximations to the
  likelihood gradient.
\newblock In {\em 25th International Conference on Machine learning}, 2008.

\bibitem{uversky2014panning}
A.~Uversky, D.~Ramljak, V.~Radosavljevi{\'c}, K.~Ristovski, and
  Z.~Obradovi{\'c}.
\newblock Panning for gold: using variograms to select useful connections in a
  temporal multigraph setting.
\newblock {\em Social Network Analysis and Mining}, 4(1):1--13, 2014.

\bibitem{van2009dimensionality}
L.~J. van~der Maaten, E.~O. Postma, and H.~J. van~den Herik.
\newblock Dimensionality reduction: A comparative review.
\newblock {\em Journal of Machine Learning Research}, 10(1-41):66--71, 2009.

\bibitem{vincent2008extracting}
P.~Vincent, H.~Larochelle, Y.~Bengio, and P.-A. Manzagol.
\newblock Extracting and composing robust features with denoising autoencoders.
\newblock In {\em Proceedings of the 25th international conference on Machine
  learning}, pages 1096--1103, 2008.

\bibitem{wytock2013}
M.~Wytock and Z.~Kolter.
\newblock Sparse gaussian conditional random fields: Algorithms, theory, and
  application to energy forecasting.
\newblock In {\em Proc. of the 30th International Conference on Machine
  Learning (ICML-13)}, pages 1265--1273, 2013.

\bibitem{yi2003effects}
Y.~Yi and H.~Jeon.
\newblock Effects of loyalty programs on value perception, program loyalty, and
  brand loyalty.
\newblock {\em Journal of the Academy of Marketing Science}, 31(3):229--240,
  2003.

\bibitem{zhou2014}
J.~Zhou, F.~Wang, J.~Hu, and J.~Ye.
\newblock From micro to macro: data driven phenotyping by densification of
  longitudinal electronic medical records.
\newblock In {\em 20th ACM SIGKDD Internation Conference on Knowledge Discovery
  and Data Mining}, pages 135--144. ACM, 2014.

\end{thebibliography}

\end{document}